\documentclass[10pt,twocolumn,letterpaper]{article}
\usepackage[dvipsnames]{xcolor}

\usepackage[pagenumbers]{cvpr} 
\definecolor{cvprblue}{rgb}{0.21,0.49,0.74}
\usepackage[pagebackref,breaklinks,colorlinks,allcolors=cvprblue]{hyperref}
\usepackage{graphicx} 
\usepackage[dvipsnames]{xcolor}
\usepackage{booktabs}
\usepackage[normalem]{ulem}
\useunder{\uline}{\ul}{}
\usepackage{mathtools}
\usepackage{amsmath}
\usepackage{amssymb}
\usepackage{amsthm}
\usepackage{caption}
\usepackage{subcaption}
\usepackage{multirow}

\newtheorem{proposition}{Proposition}

\def\paperID{12866} 
\def\confName{CVPR}
\def\confYear{2026}

\title{Drainage: A Unifying Framework for Addressing Class Uncertainty}

\author{
Yasser Taha$^{1}$\hspace{5.5em}
Grégoire Montavon$^{2,3,*}$\hspace{5.5em}
Nils Körber$^{1,*}$\\[-0.3em]
{\tt\small tahay@rki.de \hspace{3.5em} gregoire.montavon@charite.de \hspace{3.5em} koerbern@rki.de}\\[0.8em]
{\itshape\small
$^{1}$Centre for Artificial Intelligence in Public Health Research, Robert Koch Institute, 13353 Berlin, Germany}\\
{\itshape\small $^{2}$BIFOLD\,--\,Berlin Institute for the Foundations of Learning and Data, 10587 Berlin, Germany}\\
{\itshape\small$^{3}$Institute for AI in Medicine, Charité\,--\,Universitätsmedizin Berlin, 10115 Berlin, Germany}
}

\begin{document}

\maketitle
\renewcommand{\thefootnote}{\fnsymbol{footnote}}
\footnotetext[1]{Equal Supervision.}
\begin{abstract}
Modern deep learning faces significant challenges with noisy labels, class ambiguity, as well as the need to robustly reject out-of-distribution or corrupted samples. In this work, we propose a unified framework based on the concept of a ``drainage node'' which we add at the output of the network. The node serves to reallocate probability mass toward uncertainty, while preserving desirable properties such as end-to-end training and differentiability. This mechanism provides a natural escape route for highly ambiguous, anomalous, or noisy samples, particularly relevant for instance-dependent and asymmetric label noise. In systematic experiments involving the addition of varying proportions of instance-dependent noise or asymmetric noise to CIFAR-10/100 labels, our drainage formulation achieves an accuracy increase of up to 9\% over existing approaches in the high-noise regime. Our results on real-world datasets, such as mini-WebVision, mini-ImageNet and Clothing-1M, match or surpass existing state-of-the-art methods. Qualitative analysis reveals a denoising effect, where the drainage neuron consistently absorbs corrupt, mislabeled, or outlier data, leading to more stable decision boundaries. Furthermore, our drainage formulation enables applications well beyond classification, with immediate benefits for web-scale, semi-supervised dataset cleaning, and open-set applications.
\end{abstract}

\section{Introduction}

The most common scheme for data classification, shared by most popular classification datasets and models (e.g.\ ResNet, ViT, etc.) consists of a predefined set of classes and a corresponding structure at the output of the ML model, namely, a discrete probability distribution over these classes. These models are then trained to maximize the agreement between predictions and targets, via loss functions such as the cross-entropy.

In this paper, we explore the limits of this classical approach in dealing with real-world properties of the data, such as samples that do not adhere to a fixed scheme of classes, or when the labeling process is subject to a variety of hindrances. These include the case of samples that lie at the boundary between two classes and for which the labeler is forced to make an arbitrary choice, or samples that are outright misclassified due to a labeler's inattention, imperfect domain knowledge, or automated labeling.

\begin{figure*}[ht!]
  \centering
  \includegraphics[width=\linewidth]{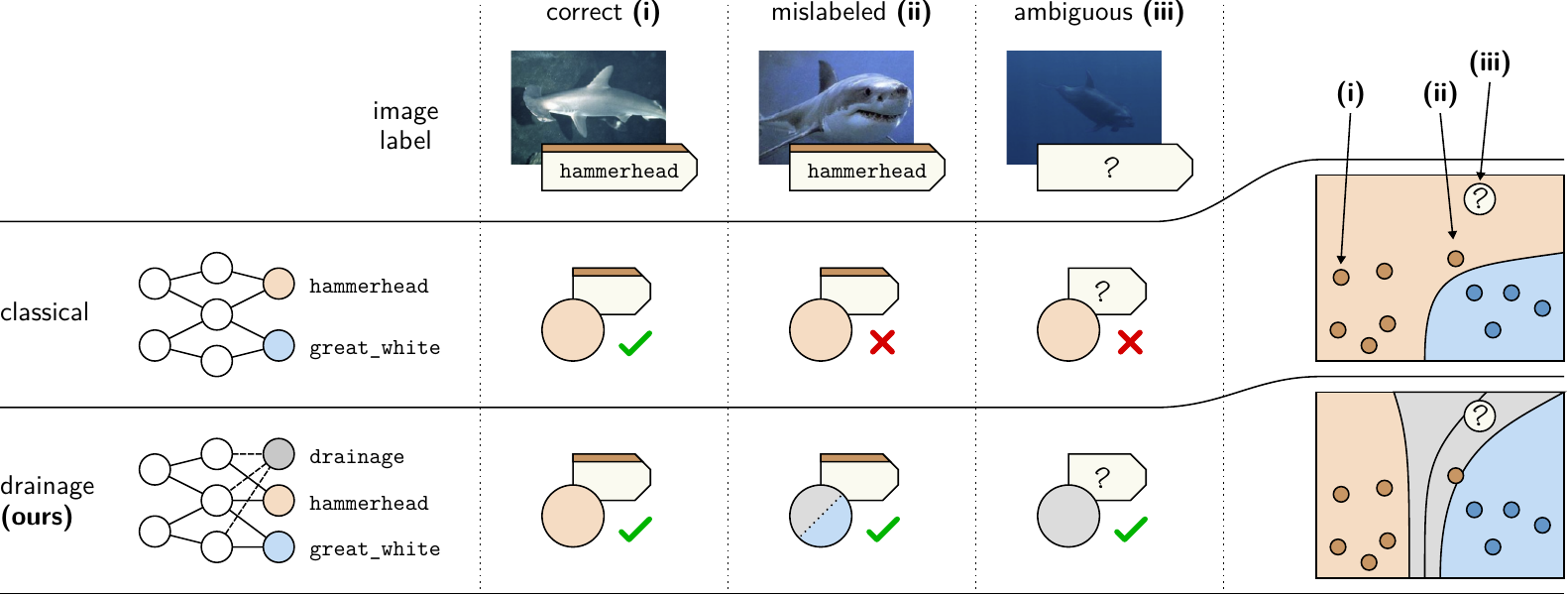}
  \caption{Different sources of class uncertainty, and the way they are handled by a classical classification model (e.g.\ softmax / cross-entropy) and by our proposed drainage model. Our drainage-based approach is more robust to mislabelings, and allows ambiguous and outlier instances to be predicted as `drainage' rather than classified arbitrarily.}
  \label{fig:overview}
\end{figure*}

To address these limitations, we propose a new model for neural network outputs, where we supplement the class-specific logits with an additional `drainage node' (illustrated in Fig.\ \ref{fig:overview}), which redirects data points that are subject to label uncertainty, whether it come from a too restrictive classification scheme, intrinsic class ambiguity in the data, or incorrect labels. We then propose a `drainage loss', which funnels these various uncertainty sources to the drainage node in a principled way.

We perform a systematic evaluation on classical image recognition datasets such as CIFAR-10, CIFAR-100, Webvision, ILSVRC12, and Clothing1M, all of which either showed a degree of uncertainty regarding their labels or were synthetically altered accordingly. Our results show significantly higher accuracy than both the cross-entropy baseline and a number of recently proposed noise-aware loss functions at high noise rates. These benchmark scores, as well as additional qualitative results emphasize the adaptability of our method to different datasets and problem formulations. Our method resolutely models specific data points as carriers of uncertainty, enabling focus on the genuine information-rich samples in the data.
We regard our contribution as a general and actionable advancement with potential impact across multiple areas of machine learning. To maintain focus within the scope of this work, we ground our study in the context of robust loss design. Nevertheless, we also illustrate the broader applicability of our approach by exploring its relevance to a proximal domain; Open Set Recognition (OSR) and demonstrate how our method can effectively complement established baselines in this research area.

Our code is available at \url{https://github.com/ZKI-PH-ImageAnalysis/Drainage}

\section{Related work}

We organize the related work section into three main themes based on conceptual relevance to our method. First, we review noise-robust learning methods, including label correction, loss correction, and robust loss functions such as APL and ANL, highlighting their strengths and limitations in handling noisy supervision.
Second, we discuss connections to heteroskedastic regression, where models explicitly account for varying uncertainty across samples; an idea conceptually aligned with our approach to handling ambiguous or noisy instances.
Finally, as our empirical observations indicate that the drainage node response is related to class ambiguity, we note a conceptual link to Open Set Recognition and briefly review relevant OSR literature.

\subsection{Robustness to Label Noise}

Learning under noisy supervision is typically addressed via three main strategies: (1) \emph{Label correction} approaches seek to improve data quality by relabeling ambiguous or mislabeled samples, often leveraging meta-learning by monitoring and influencing the label distribution \cite{MLC,noise_poison,true_label_estimate}. (2) \emph{Loss correction} methods adapt the loss function during training, weighting clean and noisy samples differently based on such estimates (e.g., \cite{mentornet, dividemix}). However, both approaches often require access to additional clean data or introduce significant computational overhead for estimating the noise model. More closely related to our approach, (3) \textit{Noise-robust loss functions} can take the form of modifying the Cross-Entropy (CE) loss or combining it with other loss function such as mean absolute error (MAE) to balance robustness and efficiency. Examples include Generalized Cross Entropy \cite{GCE} or Symmetric Cross Entropy \cite{SCE}. The Active Passive Loss (APL) framework \cite{APL} groups robust losses into “Active” and “Passive” components and combines them. Asymmetric Loss Functions (ALF) \cite{AFL}  adapt loss components to manage asymmetric noise, while ANL notably replaces APL's passive component by normalized negative loss functions (NNLFs) for improved handling of complementary labels.

Despite these advances, most robust loss functions, including ANL, apply a uniform penalty to uncertain or mislabeled samples, preventing the network from "escaping" the enforced loss in areas of high uncertainty. This limitation persists regardless of the specific robust loss design, hindering optimal learning when faced with ambiguous, outlier, or instance-dependent noise.

Our drainage loss framework addresses this core limitation, enabling neural networks to dynamically redirect ambiguous samples and escape uniform penalties in uncertain regions; a capability not provided by existing robust loss functions. 

\subsection{Heteroskedastic Models}

Other related works include `heteroskedastic' approaches, which augment predictions with additional output neurons representing predictive uncertainty \cite{nix94,bishop94,DBLP:conf/nips/Lakshminarayanan17}. In such approaches, the task is typically formulated as learning an input-conditioned output distribution, e.g.\ $y|x \sim \mathcal{N}(\mu(x),\sigma(x))$ via maximum likelihood, where $\mu$ and $\sigma$ are the two neural network outputs representing the noise and the uncertainty, respectively. Our approach relates to it by the presence of a drainage node, which can be interpreted as an uncertainty model. With the exception of a few works (e.g.\ \cite{DBLP:conf/nips/MalininG18}), the field of heteroskedastic modeling has so far mainly focused on regression tasks.

\subsection{Open Set Recognition}

Because our approach is capable of allocating probability away from the given classes towards a drainage node, our method connects to broader sets of literature, including open set recognition (OSR).
OSR approaches fall broadly into two categories: those that threshold the maximum predicted class probability to reject unknowns  \cite{osr_reject_or_not,osr_svm_reject,osr_better_mls,osr_lecun}, and those that estimate an explicit uncertainty or background score \cite{osr_openmax, osr_counterfactual}. Our method falls into the second category, where the drainage neuron produces a dedicated output probability that acts as an explicit uncertainty or "unknown" score. This allows the model to assign samples with ambiguous, out-of-distribution, or noisy characteristics directly to the drainage node, facilitating principled rejection without relying solely on thresholds over known class probabilities. This approach aligns with background or open-set score estimation methods such as OpenMax \cite{osr_openmax} and counterfactual uncertainty \cite{osr_counterfactual}, but differs by integrating the uncertainty routing within the model architecture and loss function, enabling end-to-end training for both classification and open-set recognition.

Softmax-based baselines, such as Maximum Softmax Probability (MSP), remain commonly used in OSR research \cite{osr_lr}, with recent advances refining these baselines via improved image classification techniques \cite{osr_better_mls}. We adopt MSP as our baseline for direct comparison.


\section{The Drainage Model}

In this section, we present our proposed `Drainage' framework, which extends the family of cross-entropy losses to address mislabeling and class uncertainty. We consider the classical setup of a fixed data scheme where each data point comes from a given input domain and with a fixed set of classes $1, \dots, C$. We assume this data scheme is reflected in the underlying ML architecture, where for each class $i$ we have one associated output neuron encoding the class logit $z_i$.

The starting point of our method is to include an additional output to the ML model, the `drainage node' (shown in Fig.\ \ref{fig:overview}) with associated logit $z_d$, meaning the set of output neurons becomes
\begin{align}
\big\{ (z_i)_{i=1}^C , z_d \big\}.
\end{align}
This collection of logits can be passed to a softmax function to produce a probability distribution over the output neurons. The drainage node serves to incorporate any excess probability that cannot be reasonably be allocated to the different classes, e.g.\ due to class ambiguity caused by lack of visual evidence (cf.\ Fig.\ \ref{fig:overview}, column (iii)). At inference time, we distinguish between two operating modes. First, `\textit{open drainage}', where the model assigns probability to all output neurons:
\begin{align}
\forall_{i=1}^C:~p_i = \frac{\exp(z_i)}{\mathcal{Z}} ~~, ~~ p_d = \frac{\exp(z_d)}{\mathcal{Z}}~,
\end{align}
with $\mathcal{Z} = \exp(z_d) + \sum_{i} \exp(z_{i})$, and second, `\textit{closed drainage}', where the prediction is constrained to be among the fixed set of classes, i.e.\
\begin{align}
\forall_{i=1}^C:~p_i = \frac{\exp(z_i)}{\mathcal{Z}^\ast}
\end{align}
with $\mathcal{Z}^\ast = \sum_{i} \exp(z_{i})$. While closed drainage formulation serves as a drop-in solution for classification, the open drainage formulation informs the training process, specifically can be passed to a novel loss function which we present below.

\begin{figure*}[t!]
\centering
\includegraphics[width=\textwidth]{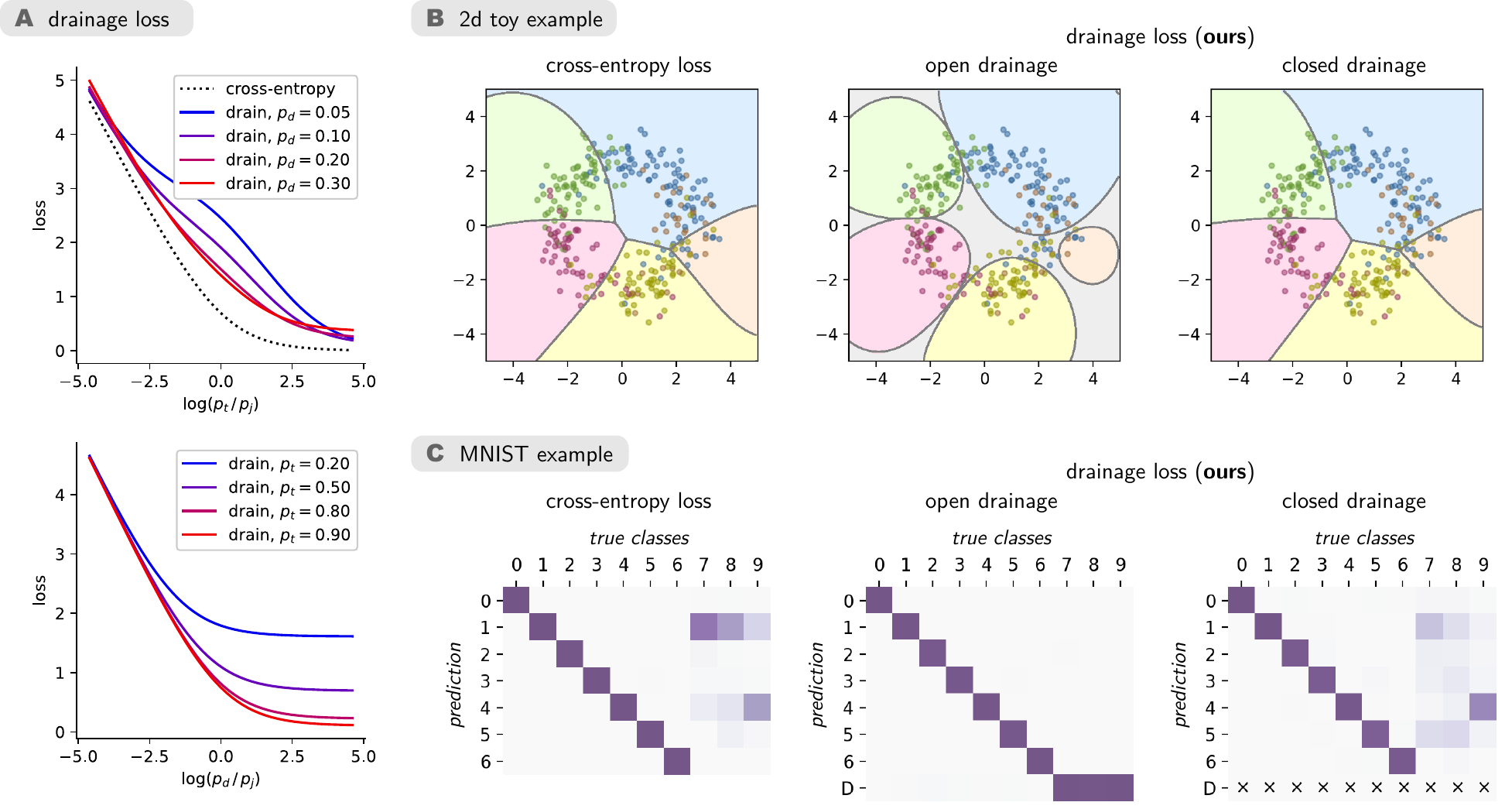}
\caption{\textbf{A.} Analysis of the drainage loss ($\alpha,\beta=1$) under different drainage levels $p_d$ and target allocation levels $p_t$, and exhibiting the monotonicity properties predicted in Propositions \ref{proposition:pjpt} and \ref{proposition:pjpd}. \textbf{B.} Application of our method on a two-dimensional toy example, where we observe the emergence of drainage dominated regions (in gray), and the effect drainage has in refining the decision boundaries between classes. \textbf{C.} Application of our method to the MNIST data. Here, we randomly relabel all training instances of classes 7,\,8,\,9 to labels 0 to 6. This causes instances from classes 7,\,8,\,9 to be systematically predicted as drainage on the test set.}
\label{fig:toy}
\end{figure*}

\subsection{Drainage Loss} We propose a novel loss function that leverages the drainage node. Our drainage loss is designed to encourage instances with limited evidence for the target class to activate the drainage node rather than nodes representing non-target classes. Consider an instance labeled to be of class $t$, denote by $\mathcal{J} = \{1,\dots,C\} \setminus \{t\}$ the set of non-target classes with $p_\mathcal{J} = \sum_{j \in \mathcal{J}}p_j$ the probability associated to them, and recall that $d$ is an index for the drainage node. We define the drainage loss as:
\begin{align}
\ell(p,t) &= \log \Big( 1 + \alpha \cdot \Big(\frac{p_d}{p_t} + \frac{p_\mathcal{J}}{p_t}\Big) + \beta \cdot \frac{p_\mathcal{J}}{p_d}\Big)
\label{eq:drainage-prob}
\end{align}
where $\alpha,\beta > 0$ are hyperparameters. The same loss function is also expressible in terms of logits as:
\begin{align}
\ell(p,t) = \mathrm{LSE} \big\{0,~
&z_d-z_t+\log \alpha, \nonumber\\
(&z_j-z_t+\log \alpha)_{j \in \mathcal{J}}, \nonumber\\
(&z_j-z_d + \log \beta)_{j \in \mathcal{J}} \big\}
\label{eq:drainage-lse}
\end{align}
In this equation, `LSE' denotes a log-sum-exp function, interpretable as a soft max-pooling. Minimizing a LSE requires that all terms in the pool are small. This can be interpreted as enforcing a set of soft-constraints on the target, drainage, and non-target logits. In particular, evidences for the drainage $z_d$ and for all of non-target classes $z_j$ are encouraged not to exceed the target evidence $z_t$, and more interestingly, the drainage logit $z_d$ is encouraged to surpass non-target logits $z_j$. The loss function is plotted in panel A of Fig.\ \ref{fig:toy}, and its behavior on two toy examples (described in more details in
Supplementary Note A%
) is shown in panels B and C.

The hyperparameters $\alpha$ and $\beta$ in the drainage loss control the level of enforcement of the multiple soft-constraints. For example, setting $\beta > \alpha$ puts the emphasis on the drainage node surpassing the incorrect classes, which result in an expansion of the regions predicted as drainage.

\paragraph{Relation to Cross-Entropy} It is useful to compare the proposed drainage loss to the classical cross-entropy loss. The latter is usually expressed compactly as `$-\log p_t$' but it can also be developed as:
\begin{align}
\ell_\text{CE}(p,t)
&= \log \Big( 1 + \frac{1-p_t}{p_t} \Big)\\
&= \mathrm{LSE}\big\{ 0, 
(\big.z_j-z_t)_{j \in \mathcal{J}}
\big\}
\label{eq:ce}
\end{align}
where we observe a similar pooling structure to that of the drainage loss. In particular, if we choose $\alpha=1$ and $\beta \to 0$ in Eq.\ \eqref{eq:drainage-prob}, and set the drainage node to its optimum under these specific parameters (i.e.\ $p_d=0$), the drainage loss reduces to the cross-entropy loss.

\paragraph{Theoretical Analysis} Here, we present a set of desirable monotonicity and convexity properties satisfied by the drainage loss.
\begin{proposition} Any reallocation of probability from non-target to target reduces the drainage loss, specifically, $\ell(p_t+\delta,p_d,p_\mathcal{J}-\delta) \leq \ell(p_t,p_d,p_\mathcal{J})$ for any probability vector $(p_t,p_d,p_\mathcal{J})$ and perturbation $0 \leq \delta \leq p_\mathcal{J}$.
\label{proposition:pjpt}
\end{proposition}
\begin{proposition}  Any reallocation of probability from non-target to the drainage node reduces the loss, specifically, $\ell(p_t,p_d+\delta,p_\mathcal{J}-\delta) \leq \ell(p_t,p_d,p_\mathcal{J})$ for any probability vector $(p_t,p_d,p_\mathcal{J})$ and perturbation $0 \leq \delta \leq p_\mathcal{J}$.
\label{proposition:pjpd}
\end{proposition}
\noindent Proofs are given in 
Supplementary Note B%
. The first property ensures, under constant $p_d$, consistent behavior with non-drainage loss functions such as the cross-entropy. The second property ensures a preference for uncertainty over incorrect predictions. Both monotonicity properties can also be observed in panel A of Fig.\ \ref{fig:toy}.
\begin{proposition}
    The drainage loss, expressed as a function of the logits $\{(z_i)_{i=1}^C,z_d\}$, i.e.\ as in Eq.\ \eqref{eq:drainage-lse}, is convex. 
\end{proposition}
\noindent Convexity results from log-sum-exp being convex with its input, and its input being a linear function of the logits. Likewise, application of the drainage loss to a linear model gives rise to a convex optimization problem.

\section{Experimental Setup}

\subsection{Datasets}
We conduct experiments on four benchmark datasets: CIFAR-10, CIFAR-100, WebVision, ILSVRC 2012 validation set, and Clothing-1M.
The CIFAR-10 dataset \cite{cifar10-100} consists of 60,000 images across 10 classes. We use 50,000 samples with varying label noise level for training and 10,000 clean samples for testing. The CIFAR-100 dataset \cite{cifar10-100} includes 100 classes grouped into 20 super-classes, with 600 images per class. Similar to CIFAR-10, 50,000 noisy labeled samples are used for training and 10,000 for evaluation.
The WebVision dataset \cite{webvision} contains over 2.4 million web images collected using queries derived from the 1,000 class labels of the ILSVRC 2012 benchmark \cite{imagenet}. Following the “Mini” setting from \cite{mentornet}, we use the first 50 classes of the Google-resized subset for training. Evaluation is performed on the same 50 classes from both the ILSVRC 2012 validation set and the WebVision validation set, which serve as clean evaluation data.
The Clothing-1M dataset \cite{clothing1m} is a large-scale collection of real-world clothing images from online shopping websites, comprising 14 categories and approximately 1 million training samples with about 40\% label noise. It also provides a clean subset divided into a training set (50k samples), a validation set (14k samples), and a test set (10k samples).
\subsection{Noise Generation}
Noise Generation:
We follow standard practices for generating label noise in image classification experiments using the CIFAR-10 and CIFAR-100 datasets.
For class-dependent noise, noisy labels are created according to established approaches in prior works \cite{APL, AFL, loss_correction}.
For CIFAR-10, flipping `{truck}'\,$\to$\,`{automobile}', `{bird}'\,$\to$\,`{airplane}, `{deer}'\,$\to$\,`{horse}', `{cat}'\,$\to$\,`{dog}'.  For CIFAR-100, the 100 classes are grouped into 20 super-classes with each has 5 sub-classes, and each class are flipped within the same super-class into the next in a circular fashion.
The noise rate $\mu$ is varied as 
$\mu\in\{0.2,0.3,0.4,0,45\}$.
For instance-dependent noise, we adopt the part-dependent noise (PDN) model from \cite{PDN}, with noise rates 
$\mu\in\{0.2,0.4,0.5\}$, where noise is synthesized based on the prediction errors of a deep neural network.

For real-world noise, we utilize the “Worst” label set of CIFAR-10N and the “Noisy” label set of CIFAR-100N \cite{noisy_cifar}, which contain human-annotated labels exhibiting natural annotation errors.

\subsection{Baseline methods}

For CIFAR-10 and CIFAR-100 under asymmetric and instance-dependent noise settings, we evaluate several baseline methods, including state-of-the-art approaches: (a) Generalized Cross Entropy (GCE) \cite{GCE}, (b) Symmetric Cross Entropy (SCE) \cite{SCE}, (c) Active Passive Loss (APL) \cite{APL}, specifically NCE+RCE, (d) Asymmetric Loss Functions (AFL) \cite{AFL}, specifically NCE+AGCE, and (e) Active Negative Loss (ANL) \cite{ANL}. For comparison, we also train networks using the standard Cross Entropy (CE) loss.

We evaluate CE, GCE, SCE, AFL, APL, and ANL-CE (the ANL based on CE) on the CIFAR-10N, CIFAR-100N, WebVision, and ILSVRC12 datasets. For the Clothing-1M dataset, we also follow the experimental setup in \cite{ANL}, using CE, GCE, and ANL-CE as baseline methods.

\subsection{Parameter selection}

For all baseline methods, we used hyperparameters as specified in their original implementations. Consistent with \cite{ANL}, L1 regularization was found effective in improving generalization for the ANL loss but not for the APL loss, which we also confirmed for both APL and AFL losses. To provide a more comprehensive evaluation, we extended our experiments to include CE, SCE, and GCE losses with L1 regularization and observed that L1 indeed helps these losses overcome the overfitting issues noted in \cite{ANL}.

While prior work typically reports CE, SCE, and GCE with L2 regularization, we adopt the regularization strategy that yields the best performance on CIFAR-10 for each loss. Concretely, CE, GCE and SCE losses produce best results when trained with L1 regularization, whereas ALF and APL losses perform better with L2 regularization, aligning with observations from \cite{ANL}. Similar to ANL, our proposed drainage approach similarly benefits from L1 regularization. For completeness, we also report CE, GCE, and SCE results with L2 regularization, alongside APL and AFL trained with L1 regularization in Table \ref{table:results_additional}; these consistently performed worse than their best counterparts. Detailed hyperparameter settings for all methods and datasets are provided in Supplementary Table \ref{table:parameters}.

Following \cite{APL, AFL, ANL}, we use an 8-layer CNN for CIFAR-10 and a ResNet-34 \cite{resnet} for CIFAR-100. Models are trained for 120 and 200 epochs on CIFAR-10, and CIFAR-100, respectively, using SGD with momentum 0.9, cosine learning rate decay, and initial learning rates of 0.01 (CIFAR-10) and 0.1 (CIFAR-100). For drainage loss, the learning rate is reduced tenfold for CIFAR-100. Standard augmentations of random width/height shift and horizontal flip are applied to all losses.

For CIFAR-10 and CIFAR-100, we introduce asymmetric noise with a rate of 0.4 and test for $\alpha \in \{0.1, 1, 10\}$ and $L1 \in$ \{$1 \times 10^{-4}$, $2 \times 10^{-4}$\}. See Supplementary Fig.\ \ref{fig:ablation_study} for an ablation study. Following \cite{APL}, the best-performing parameters are selected using a randomly chosen validation split (20\%). The optimal values are $\alpha = 1$ for CIFAR-10 and $\alpha = 0.1$ for CIFAR-100. These configurations are then reused for all other noise variants, including asymmetric, instance-dependent, and real-world noisy datasets CIFAR-10N and CIFAR-100N.

For the WebVision dataset, we follow the experimental setup used in prior works \cite{APL,ANL,AFL}. A ResNet-50 \cite{resnet} is trained using SGD for 250 epochs with an initial learning rate of 0.4, Nesterov momentum of 0.9, and a batch size of 512. The parameters for each loss function follow their original implementations and are listed in Supplementary Table \ref{table:parameters}. The learning rate is decayed by a factor of 0.97 after every epoch. Gradient norms are clipped to 5.0 to stabilize training. All images are resized to 224×224, and standard data augmentations, including random shifts in width and height, color jittering, and horizontal flips, are applied.

For the Clothing-1M dataset, we follow the experimental protocol from previous work \cite{true_label_estimate}. The 14k and 10k clean subsets are used for validation and testing, respectively, while the 50k clean training subset is excluded. We employ a ResNet-50 \cite{resnet} pre-trained on ImageNet-V2. During preprocessing, images are resized to 256×256, mean-normalized, and center-cropped to 224×224. Training is performed using the SGD optimizer with a momentum of 0.9, weight decay of $0.001$ and a batch size of 32. We adopt the tuned parameters reported by \cite{ANL} for GCE and ANL. Parameters for each loss function are summarized in Supplementary Table \ref{table:parameters}. The network is trained for 10 epochs, with learning rates of $10^{-3}$ for the first 5 epochs and $10^{-4}$ for the remaining 5. For drainage loss, the learning rate is reduced tenfold. Standard data augmentation, including random horizontal flipping, is applied.

All results are reported across three independent runs, with mean accuracy and maximum standard deviation reported per loss. For all Drainage results, we set $\beta$ in Equation \eqref{eq:drainage-prob} to $1/\alpha$.

\section{Results}

Results of our benchmark evaluations are shown in Tables \ref{table:synthetic} and \ref{table:natural}, and demonstrate the overall higher accuracy reached by our Drainage approach across different datasets and types of label noise. We present results for each dataset in detail below.

\smallskip

\noindent \textbf{CIFAR-10~} In absence of noise or low-noise regimes, most methods are tied within one or two percentage points, with the baseline CE loss performing expectedly well, and sharing the top spots with GCE and ANL CE. Our Drainage method takes a distinct lead as noise increases. Notably, the performance gap between Drainage and the second-ranked method widens with increasing noise levels, demonstrating the overall robustness of our approach.

\smallskip

\noindent \textbf{CIFAR-100~} With its larger class space and increased label complexity, CIFAR-100 amplifies the challenges posed by noise. CE exhibits a significant performance drop compared to CIFAR-10, suggesting that its effectiveness reduces as the number of decision boundaries increases. GCE maintains consistent performance across both datasets, ranking second on instance-dependent and asymmetric noise. In contrast, ANL with CE performs well in low-noise regimes but its degrades its performance at higher noise levels. Our Drainage method outperforms all baselines, achieving accuracy margins of 9–10\% over the second-ranked method at high noise levels on both asymmetric and instance-dependent noise. This demonstrates the particular strength of our approach for handling larger label spaces under a wide range of noise conditions.

\smallskip

\noindent On \textbf{CIFAR-10N} and \textbf{CIFAR-100N}, GCE and ANL with CE achieve the highest scores on CIFAR-10N, with Drainage performing competitively. However, Drainage excels on CIFAR-100N, where noise is predominantly structured as asymmetric and instance-dependent. On \textbf{Mini \mbox{WebVision}} and \textbf{ILSVRC12 (Mini ImageNet)}, Our approach surpasses state-of-the-art results on Mini WebVision and ranks second on ILSVRC12. On  \textbf{Clothing1M}, Our model achieves around 2\% improvement over the previous state-of-the-art on this dataset, a large-scale dataset containing approximately one million training images with an estimated noise level of 40\%. Overall, these performance gains underscore the model’s robustness in handling prevalent asymmetric and instance-dependent noise typical of real-world annotation pipelines. The drainage node plays a crucial role by explicitly routing uncertain or ambiguous samples, providing a principled mechanism to adaptively manage label noise and improve generalization.

\begin{table*}
\centering
\begin{tabular}{lcccccccc}\toprule
 & CE & GCE & SCE & AFL & APL & ANL CE &  Drainage & max.\ err.\\
&&&&&&& (\textbf{ours})\\\midrule
CIFAR-10\\
~~~~~~noise $=0.0$ & \textbf{92.16} & 90.96 &  91.36 &  91.17 & 91.12 & {\ul91.76}    & 91.30 & $\pm$\textit{0.31} \\[1mm]
~~~\textit{asymmetric:}\\
~~~~~~noise $=0.2$ & 88.67 & {\ul89.09} & 86.76 &  88.59 & 88.45  & \textbf{89.34}  & 88.54 & $\pm$\textit{1.30}\\
~~~~~~noise $=0.3$ & {\ul86.53} &85.92& 81.28& 85.48 & 84.89 & 85.13 & \textbf{87.56} & $\pm$\textit{0.23}\\
~~~~~~noise $=0.4$ & {\ul82.50} & 80.64 & 74.63 & 78.19 & 77.48 & 78.07  & \textbf{84.66} & $\pm$\textit{0.59}\\
~~~~~~noise $=0.45$ & {\ul77.92} & 75.57 &71.07 & 71.74 & 72.00 & 72.53 & \textbf{82.23} &$\pm$\textit{0.77}\\[1mm]
~~~\textit{instance dependent:}\\
~~~~~~noise $=0.2$ & 85.03 & {\ul88.68} & 83.41 & 88.09 & 88.26 & \textbf{88.78} & 87.92 & $\pm$\textit{0.40}\\
~~~~~~noise $=0.4$ & 71.13 & {\ul78.43}& 64.21 & 76.29 & 76.57 & 76.54   &  \textbf{81.95} &$\pm$\textit{0.76}\\
~~~~~~noise $=0.5$ & 54.61 & {\ul59.27} & 50.30 & 48.86 & 53.62 & 58.41 & \textbf{64.22} &$\pm$\textit{0.91}\\\midrule
CIFAR-100\\
~~~~~~noise $=0.0$ & 70.05 & 64.76& 69.88&68.94& 68.19& 70.15 & \textbf{73.31} &$\pm$\textit{1.64}\\[1mm]
~~~\textit{asymmetric:}\\
~~~~~~noise $=0.2$ & 57.34 & 63.31 & 57.75& 63.72 & 62.83  & {\ul66.09}  & \textbf{70.04} & $\pm$\textit{0.81}\\
~~~~~~noise $=0.3$ & 50.81 & 60.48 & 49.30 & 56.63& 55.52 & {\ul59.85}  & \textbf{67.93} & $\pm$\textit{0.86}\\
~~~~~~noise $=0.4$ &41.63 & {\ul55.08} & 41.61 & 44.23& 42.60 & 46.01  & \textbf{61.55} &$\pm$\textit{0.13}\\
~~~~~~noise $=0.45$ & 34.81 & {\ul42.70} & 36.63 & 37.39 & 35.15 & 37.60 & \textbf{52.69} &$\pm$\textit{1.00}\\[1mm]
~~~\textit{instance dependent:}\\
~~~~~~noise $=0.2$ & 58.10 & 63.65 & 56.54 & 64.6 & 63.61 & {\ul66.20}  & \textbf{67.16} &$\pm$\textit{1.38}\\
~~~~~~noise $=0.4$ & 42.43 & {\ul57.10} & 42.50 & 52.46 & 50.28 &  56.55 & \textbf{60.63} &$\pm$\textit{0.97}\\
~~~~~~noise $=0.5$ & 35.31 & {\ul48.12} & 35.11 & 42.11 & 40.52 &  45.67 & \textbf{53.01} &$\pm$\textit{1.59}\\
\bottomrule
\end{tabular}
\caption{Results on datasets with artificially added label noise, and where noise is either added homogeneously on class members but with asymmetry between classes, or non-homogeneously across instances (instance-specific). Best and second best results are shown in bold and underline respectively. The column `max.\ err.' indicates the maximum standard error over all compared methods and serves as a significance indicator.}
\label{table:synthetic}
\end{table*}

\begin{table*}
\centering
\begin{tabular}{lccccccccccc}\toprule
 & Number &Noise& CE & GCE & SCE & AFL & APL & ANL CE  & Drainage & max.\ err.\\
&of classes&level (\%)&&&&&&& (\textbf{ours})\\\midrule

CIFAR-10N &10&40.2& 76.93   & \textbf{81.70} & 73.43 &  80.07 & 79.80 & {\ul80.54}  & 79.85 & $\pm$\textit{0.38} \\

CIFAR-100N &100&40.2&48.17 & 51.19 & 48.51 & 55.29& 54.62& {\ul56.62} & \textbf{57.77} & $\pm$\textit{1.03}\\

WebVision &50 &20.0 & 61.27 & 57.16 & 65.52 & 66.20 & 64.77 & {\ul 67.53} & \textbf{68.80} &$\pm$\textit{1.82}\\

ILSVRC12 &50&0.0& 56.32 & 55.01 & 60.82 & 62.16 & 61.16 & \textbf{65.77} & {\ul 64.90} &$\pm$\textit{1.73}\\

Clothing1M &14&38.5& 67.93 & 68.05 &67.96$^\ast$&68.94$^\ast$&68.70$^\ast$& 68.44  & \textbf{70.42} &$\pm$\textit{0.55}\\
\bottomrule
\end{tabular}
\caption{Results on real-world image datasets featuring real human labeling noise (e.g.\ resulting from mistakes or subjective assessments). Best and second-best results are in bold and underline. `max.\ err.' indicates the maximum standard error over all methods. Results marked with $^\ast$ are without fine-tuning.}
\label{table:natural}
\end{table*}

\subsection{Qualitative results}
\begin{figure*}[t!]
  \centering
      \includegraphics[width=.91\linewidth,trim=0 20 0 0]{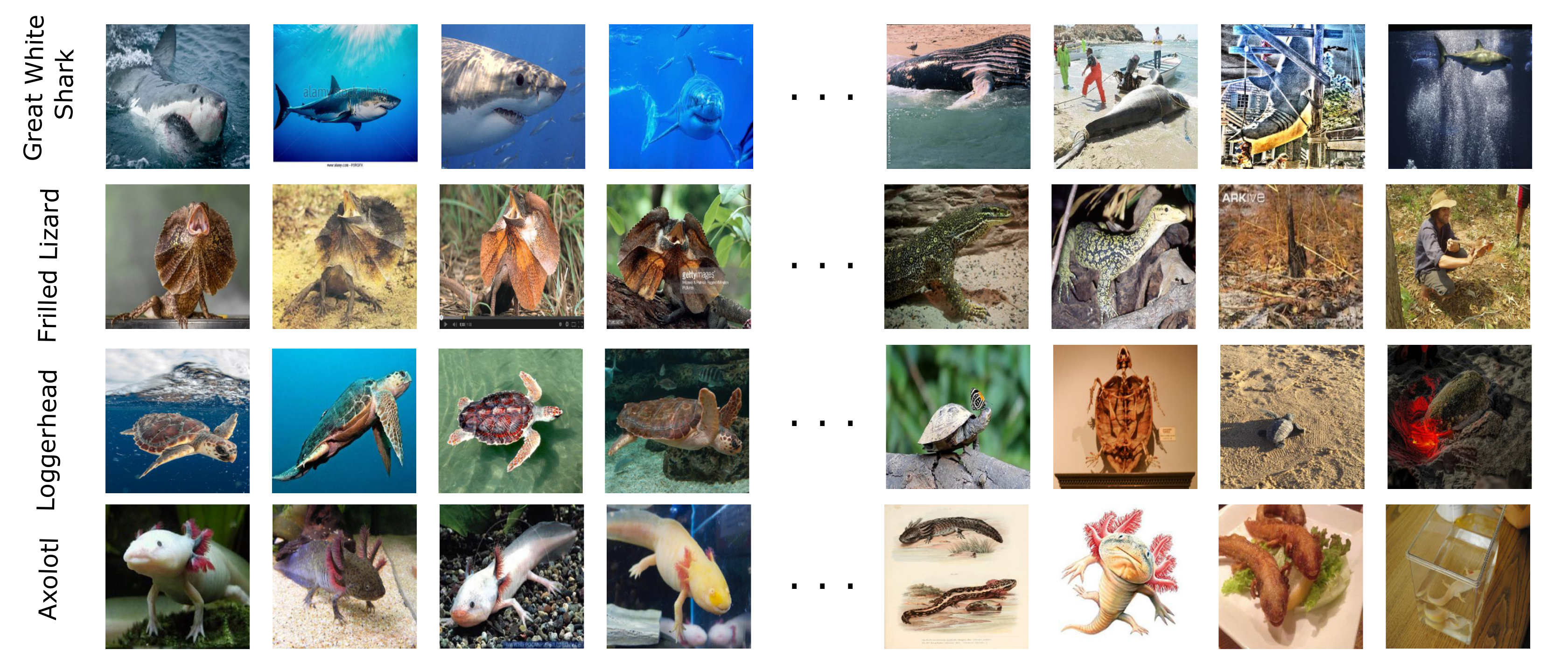}
  \caption{Examples from four Mini-WebVision \cite{mentornet} classes, with validation samples ordered by increasing drainage response $p_d$.}
  \label{fig:drain_images_webvision_validation}
\end{figure*}

In Fig.\ \ref{fig:drain_images_webvision_validation}, the effect of the drainage neuron is depicted showing multiple rows of images from the WebVision validation set \cite{webvision}, each ordered left-to-right by increasing drainage response. This ordering highlights clear trends: samples on the left are more class-representative and typical, while those towards the right become progressively ambiguous, outlier-like, or potentially mislabeled. Since the WebVision validation set is annotated via Amazon Mechanical Turk, some labeling noise remains; evident in examples like a whale labeled as a great white shark, or a lizard lacking key features mislabeled as a frilled lizard. We also observe drawings or artistic renderings that reflect covariate shift \cite{osr_lr}, further increasing uncertainty (example: Axolotl). Conducting the same analysis on the noisier training set (cf.\ Supplementary Fig.\ \ref{fig:drain_images_webvision_train}) reveals that images with the strongest drainage response are often genuinely unknown or out-of-distribution classes.

\section{Adaptation to Open Set Recognition}
\label{sec:osr_results}

\begin{figure}[h!]
\centering \footnotesize
\parbox{.45\linewidth}{\centering ~~~~~learned $z_d$}%
\parbox{.45\linewidth}{\centering ~~~~~constant $z_d$}\\
\includegraphics[width=.45\linewidth,clip,trim=8 10 0 5]{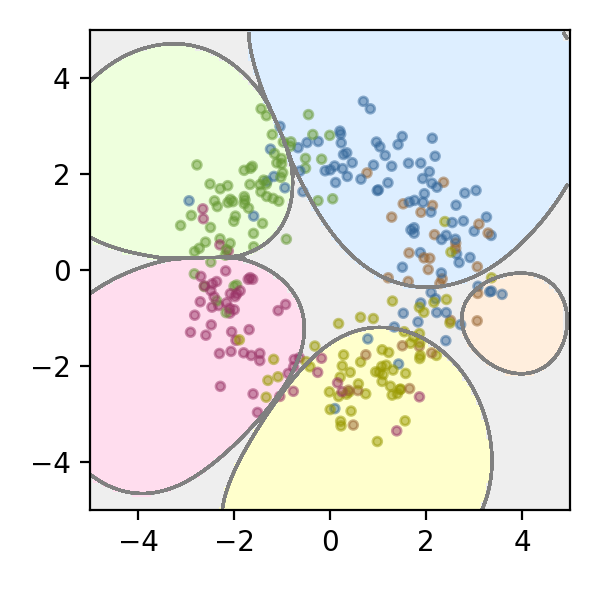}%
\includegraphics[width=.45\linewidth,clip,trim=8 10 0 5]{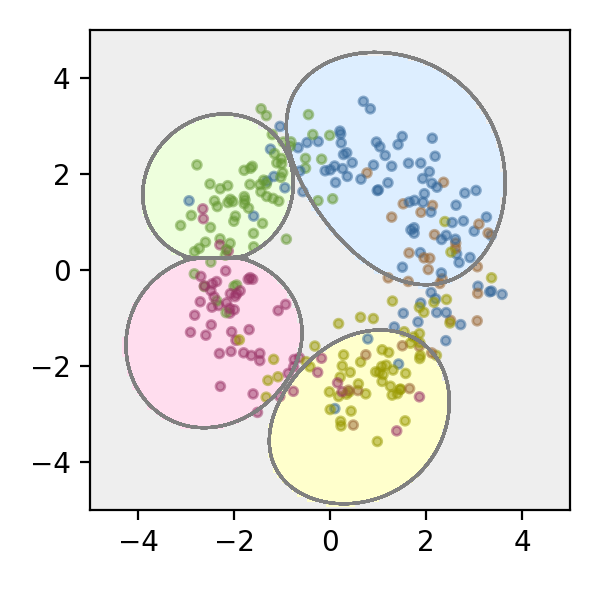}
\caption{Effect of modeling decisions for $z_d$ on our two-dimensional toy example. On the right, the original learned model $z_d$ is replaced by the constant model $z_d=2$.
}
\label{fig:toy-openset}
\end{figure}

In the benchmark evaluations of Table \ref{table:synthetic} and \ref{table:natural}, we have considered classification performance under closed drainage node. In other words, we have used the drainage node as a regularization mechanism rather than as an explicit prediction outcome. However, the insights gathered from Fig.\ \ref{fig:drain_images_webvision_validation} suggest that the drainage node can serve as a model on its own, and may be useful specifically for Open Set Recognition (OSR). Motivated by this use case, we make the modeling decision to disconnect $z_d$ from the network, i.e.\ we set $z_d = \text{cst.}$ This emphasizes the role of the drainage node in collecting all samples for which none of the class logits strongly responds, that is, samples not containing enough visual evidence for any of the existing classes. The effect of this modeling decision is illustrated for our two-dimensional toy example in Fig.\ \ref{fig:toy-openset}. We observe that $z_d$'s constant modeling helps envelop the data more tightly while maintaining boundaries between existing classes, thereby making the model amenable to OSR.

\begin{table*}
\centering
\begin{tabular}{llll} \toprule
Loss function                 & Scoring rule    & SVHN    & CIFAR-10           \\\midrule
CE & MSP ($\max_{i} p_i$)      & 91.0 \scriptsize {(\textit{91.92, 89.35, 92.08, 90.44, 91.38})}         & 67.3 \scriptsize {(\textit{70.13, 59.03, 68.79, 72.24, 66.31 })}      \\[2mm]
Drainage (\textbf{ours}) &  MSP ($\max_{i \notin \{d\}} p_i$) & \underline{92.0} \scriptsize {(\textit{91.05, 91.94, 92.18, 93.09, 91.58})} & \underline{72.2} \scriptsize {(\textit{72.68, 67.61, 72.44, 75.53, 72.53})} \\
'' & $p_d$     & \textbf{92.3} \scriptsize {(\textit{92.31, 92.16, 92.12, 92.97,91.96})}          & \textbf{72.4} \scriptsize {(\textit{72.97, 68.35, 71.97, 75.78, 73.00})}\\
\bottomrule
\end{tabular}
\caption{Evaluation on Open Set Recognition. We report the ROC AUC metric for each dataset, model and scoring rule. For the drainage approach, we use $z_d = \text{cst.}$ We show the average over 5 different class splits with 4 classes randomly removed from training set for OSR, individual results shown in parenthesis. MSP for 'maximum softmax probability'. Best and second-best results are in bold and underline. }
\label{table:osr}
\end{table*}

To test this, we follow the approach in \cite{osr_counterfactual} by removing four classes from each dataset (SVHN and CIFAR-10). The remaining samples are then split into training, validation, and test sets, with hyperparameters selected using the validation set. The best model is evaluated on the test set and on the masked classes. This procedure is repeated five times, each with a different set of masked classes. The masked classes are identical across both reported losses. We fix $\alpha$ and 
$\beta$ to 1.0 and tune 
$z_d$ over $\{1.0,2.0,3.0,4.0,5.0\}$. Our model comprises a simple 3-convolutional-layer backbone followed by two fully connected layers with ReLU activations. As baselines, we report the maximum softmax probability (MSP), alongside our proposed Drainage.

Results are shown in Table \ref{table:osr} (for closed-set accuracy, cf.\ Supplementary Table \ref{table:osr-closed}). We observe that, without further tuning, our method outperforms the cross-entropy baseline. This shows the flexibility of the drainage node approach in adapting to other problem formulations than classification.

\section{Conclusion}

We introduced Drainage, a unifying approach that allows the underlying model to separate model uncertainties from class-specific probabilities. It comprises an extra output node and a cross-entropy-derived loss function, that encourages the model to allocate probabilities to the drainage node rather than to incorrect classes. This helps the model focus on learning class-relevant features, even in very noisy conditions. We evaluated Drainage at different levels of label noise and found that it outperforms all other methods by a large margin in conditions involving high noise levels. When applied to a real-world setup using noisy data, the model demonstrates its ability to learn class boundaries while retaining class ambiguity in the drainage node. Furthermore, we demonstrate that our approach can be used for related tasks, such as Open Set Recognition, providing a performance boost  over the MSP baseline by either incorporating it within MSP or simply using the drainage node probability directly.

\textbf{Limitation~} Although Drainage performs effectively across diverse settings, it requires tailored adaptation depending on the specific application, such as noise robust classification versus OSR.

\bibliographystyle{plain}
\bibliography{references} 

\clearpage

\onecolumn

\vspace*{5mm}

\begin{center}
\Large {\bf
Drainage: A Unifying Framework for Addressing Class Uncertainty}\\[2mm]
\large \textsc{(Supplementary Material)}
\end{center}

\bigskip

\renewcommand{\thesection}{Supplementary Note \Alph{section}}
\setcounter{section}{0}
\setcounter{table}{0}
\setcounter{figure}{0}
\captionsetup[table]{name=Supplementary Table}
\captionsetup[figure]{name=Supplementary Figure}
\setcounter{page}{1}

\section{Details of the Toy Experiment}
\label{section:toy-details}

In this note, we present details of the experiments presented in the main paper in Fig.\ \ref{fig:toy} (panels B and C).

\paragraph{2d Toy Experiment} We build a dataset where data points are placed on a ring centered at the origin. Each data point's radius is drawn from a Gaussian distribution and its angle is class-dependent. Points of the first class (depicted in brown in Fig.\ \ref{fig:toy}) are relabeled with probability $0.5$ to the second class (shown in blue). To classify the data, we consider a simple linear classification model $W \Phi(x)$. The function $\Phi:\mathbb{R}^2 \to \mathbb{R}^{|\mathcal{Q}|}$ is a fixed radial basis expansion defined as:
$\Phi(x) = (\exp(-\gamma \|x-q\|^2))_{q \in \mathcal{Q}}$ where the set $\mathcal{Q}$ defines $400$ basis centers, arranged on a $20\times20$ mesh grid spanning the input domain $[-5,5]^2$, and where we set $\gamma=0.25$. The parameter $W$ is a matrix of size $(C+1) \times 400$, learned from the data, and mapping the $400$-dimensional radially expanded inputs to the logits for the $C$ classes and the drainage node. This one-layer setting makes the training objective straightforward to optimize, and produces solutions that do not depend on the training parameters. We train the model until convergence and regularize the model by setting the weight decay parameter to $0.1$.

\paragraph{MNIST Toy Experiment}
We employ the network used for our Open-Set Recognition task described in Section \ref{sec:osr_results}. The MNIST dataset is split into training and test sets. We randomly relabel classes 7, 8, and 9 to labels 0–6 with uniform probability, simulating missing class instances in the training data. The test set labels are kept unchanged. We evaluate the trained model under two conditions: with closed drainage $\max_{i} p_i$  and without open drainage $\max_{i \notin \{d\} p_i}$ . Reported confusion matrices are presented in Figure \ref{fig:toy}.

\section{Proofs of Propositions \ref{proposition:pjpt} and \ref{proposition:pjpd}}
\label{section:proofs}

Here, we give detailed proofs for the first two propositions stated in the main paper. Let us first express the drainage loss in Eq.\ \eqref{eq:drainage-prob} as the following composition of functions:
\begin{align}
\ell(u,v) = \log \big( 1 + \alpha u + \beta v \big), \qquad
u(p) =  \frac{p_d+p_\mathcal{J}}{p_t}, \qquad
v(p) = \frac{p_\mathcal{J}}{p_d}.
\end{align}
The first function monotonically increasing with $u$ and $v$. The remaining two functions will be reparameterized by a real value $s$ to model reallocation across the different probability terms. To model the reallocation from $p_\mathcal{J}$ to $p_t$, as considered in Proposition \ref{proposition:pjpt}, we reparameterize the probability terms as:
\begin{align}
\begin{pmatrix}p_t\\p_d\\p_\mathcal{J}\end{pmatrix} =
\begin{pmatrix}s\\p_d\\1-p_d-s\end{pmatrix}
\end{align}
with $0 \leq s \leq 1-p_d$, and $p_d$ treated as constant. This gives us the expressions
\begin{align}
u(s) &= \frac{1-s}{s}\\
v(s) &= \frac{1-p_d-s}{p_d}
\end{align}
which are both decreasing functions of $s$. Application of composition rules for monotonic functions leads to the observation that $\ell(s)$ is monotonically decreasing, in other words, the proposed drainage loss decreases when reallocating probability from non-target classes to the target class. We consider now the reallocation from $p_\mathcal{J}$ to $p_d$ studied in Proposition \ref{proposition:pjpd}, and consider for that purpose the reparameterization
\begin{align}
\begin{pmatrix}p_t\\p_d\\p_\mathcal{J}\end{pmatrix} =
\begin{pmatrix}p_t\\s\\1-p_t-s\end{pmatrix}
\end{align}
with $0 \leq s \leq p_t$, and $p_t$ treated as constant. This gives us the expressions
\begin{align}
u(s) &= \alpha \cdot \frac{1-p_t}{p_t}\\
v(s) &= \beta \cdot \frac{1-p_t-s}{s}
\end{align}
where $u(s)$ is constant and $v(s)$ is monotonically decreasing. Here again, applying the composition rules for monotonic functions results in the observation that $\ell(s)$ is a monotonically decreasing function. In other words, any reallocation from the non-target classes to the drainage results in a decrease of the loss.

\begin{figure*}[h!]
  \centering
      \includegraphics[width=.95\linewidth]{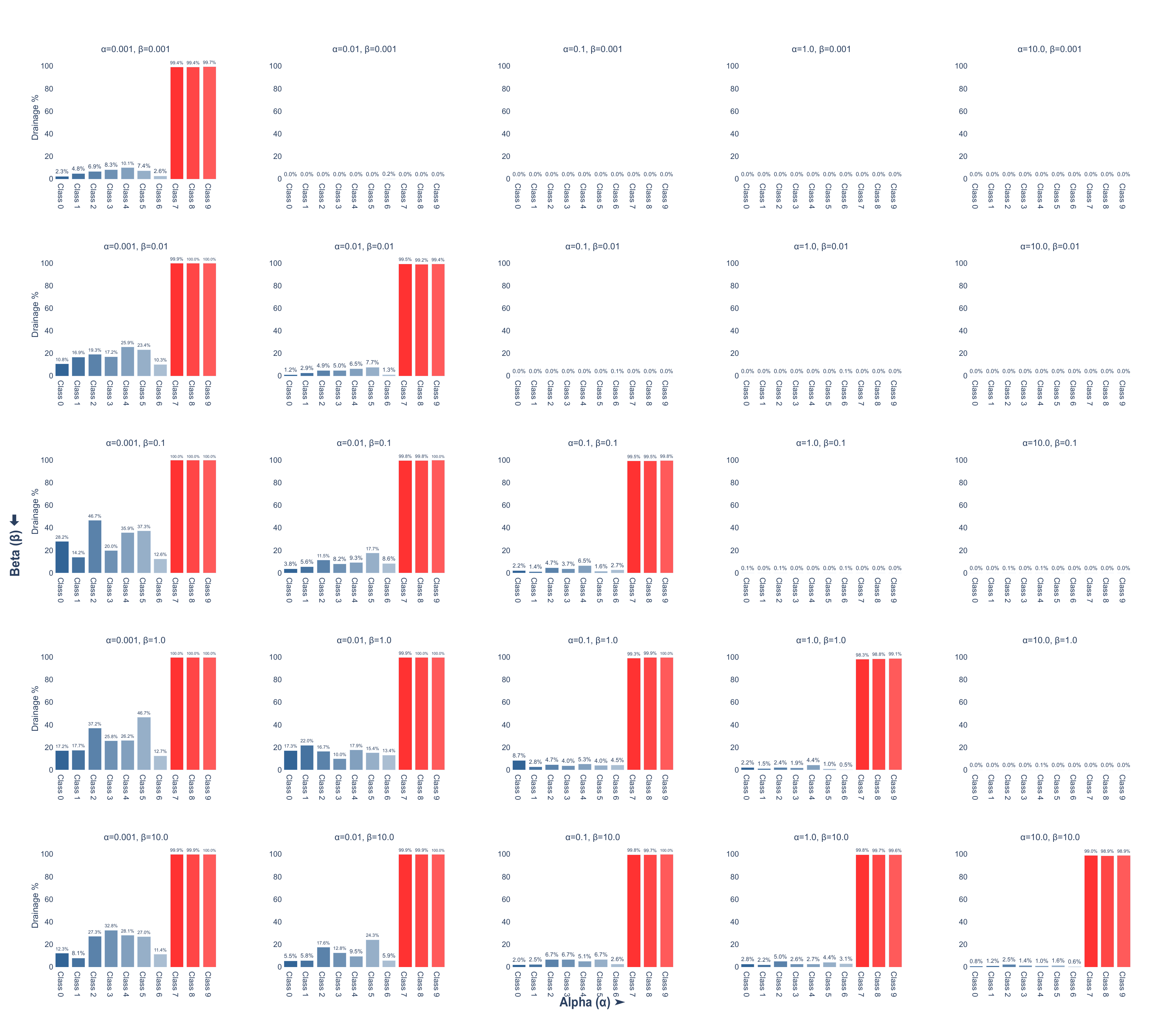}
  \caption{Effect of changing $\alpha$ and $\beta$ in  the MNIST toy example on the percentage of samples predicted as drainage per class. MNIST toy example depicted in Figure \ref{fig:toy}. In-distribution classes are in blue and Out-of-distribution in red. }
  \label{fig:training_curves}
\end{figure*}

\begin{figure*}[h!]
\centering \footnotesize

\makebox[\linewidth][c]{
\includegraphics[width=.47\textwidth]{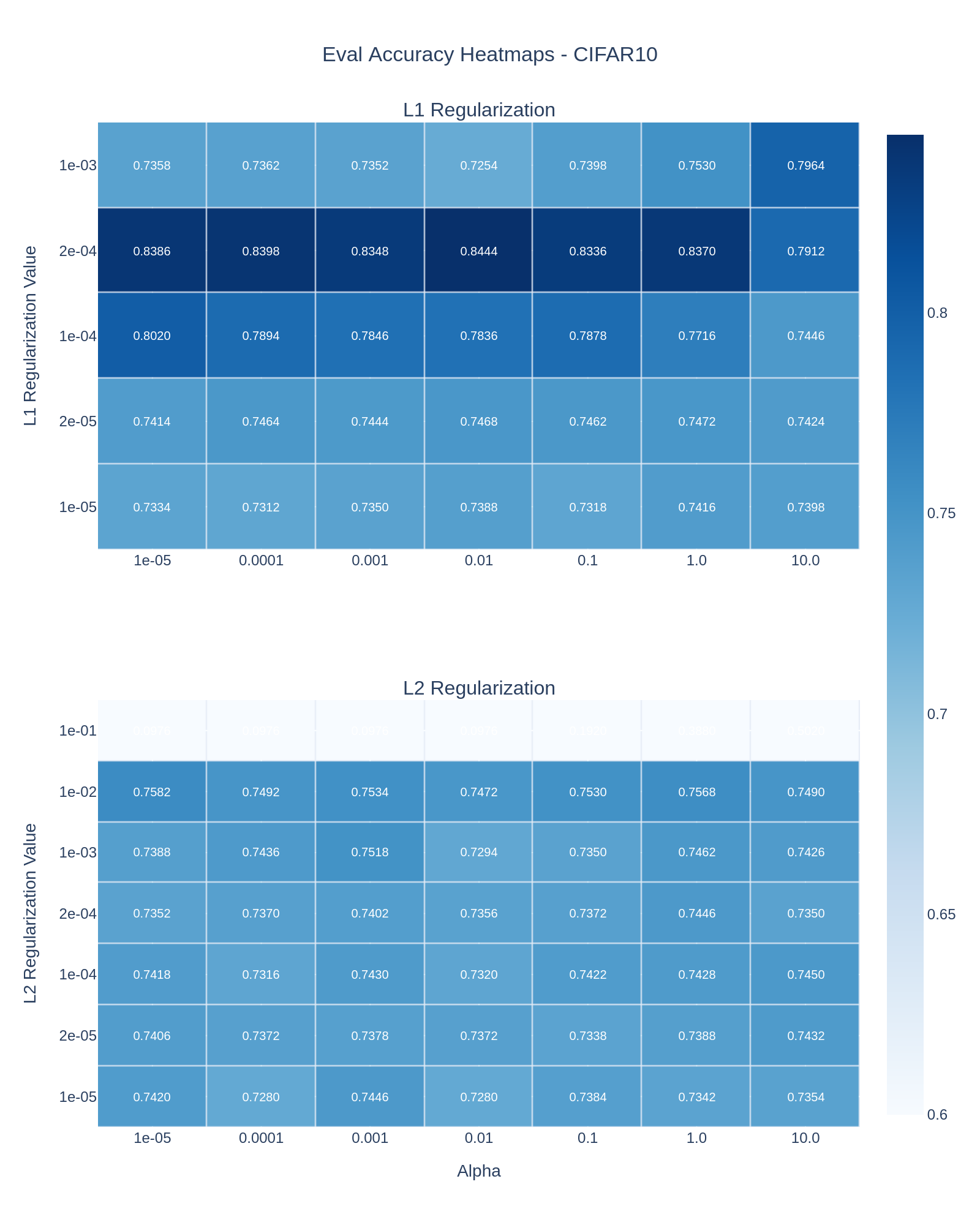}%
\includegraphics[width=.47\textwidth]{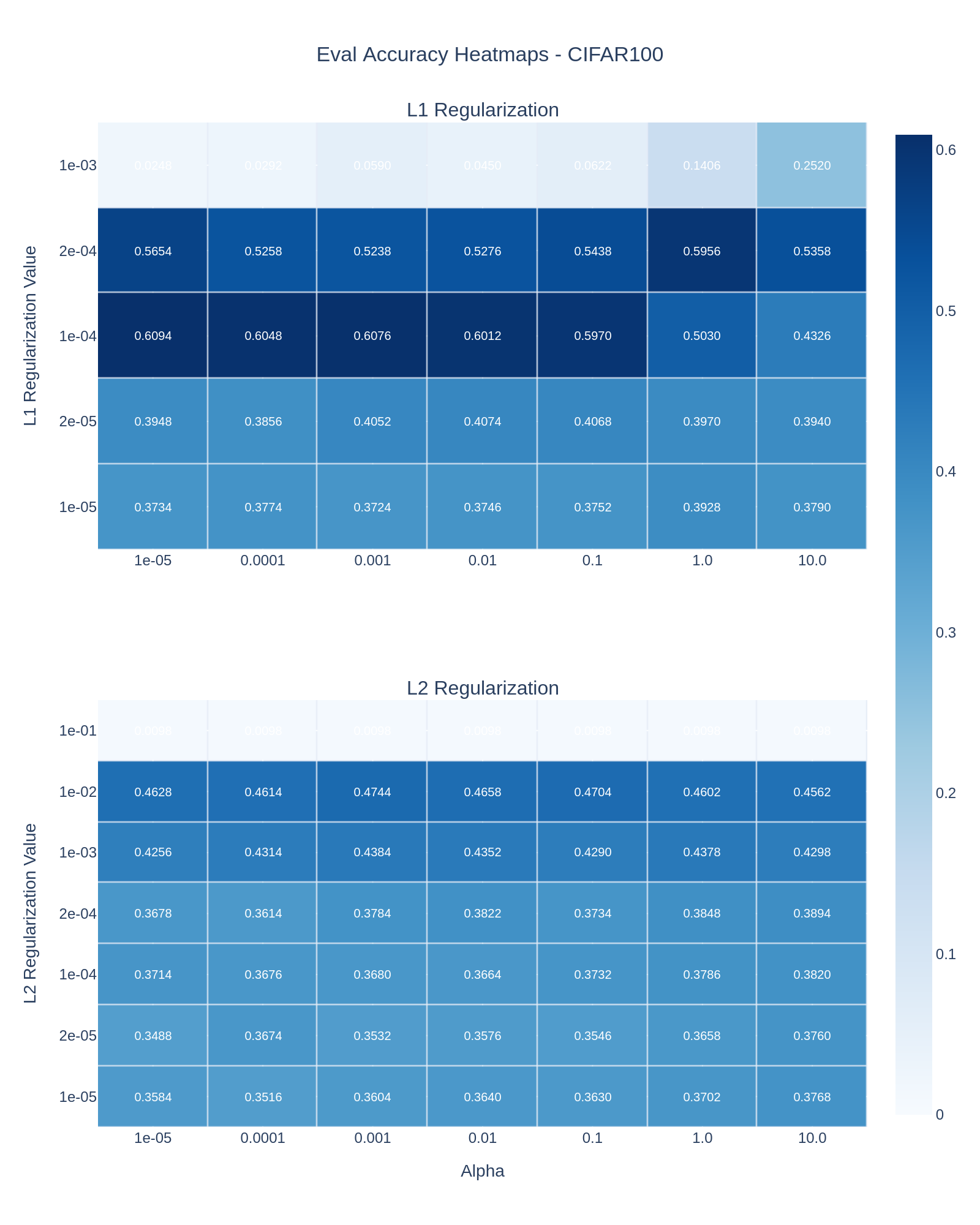}%
}
\caption{
Ablation study of the drainage loss on CIFAR-10 (left) and CIFAR-100 (right). We evaluate the effect of different regularization strengths, testing L1 and L2
coefficients of $1 \times 10^{-4}$, $2 \times 10^{-4}$, $1 \times 10^{-5}$ and $2 \times 10^{-5}$. We grid-search for the parameter $\alpha$ while setting $\beta=\alpha^{-1}$.
}
\label{fig:ablation_study}
\end{figure*}

\begin{table}[h!]
\centering \small
\makebox[\textwidth][c]{
\begin{tabular}{lcccc}\toprule
Method     & CIFAR-10              & CIFAR-100  & Webvision & Clothing1M          \\\midrule
CE($\delta$)  & L1: $1 \times 10^{-4}$& L1: $1 \times 10^{-5}$& L2: $3 \times 10^{-5}$&L1: $1 \times 10^{-3}$\\

SCE ($\alpha, \beta, \delta$)  &(0.1, 1.0, L1: $1 \times 10^{-4}$)&(6.0, 0.1, L1: $1 \times 10^{-5}$)&(10.0, 1.0, L2: $3 \times 10^{-5}$)& (5.0, 1.0, L1: $1 \times 10^{-3}$)\\

GCE ($q, \delta$) &(0.7, L1: $1 \times 10^{-4}$)&(0.7, L1: $1 \times 10^{-5}$)&(0.7, L2: $3 \times 10^{-5}$)&(0.7, L1: $1 \times 10^{-3}$)\\[2mm]

AFL ($\alpha, \beta, a, q, \delta$) &\parbox{.14\textwidth}{\centering (1.0, 4.0, 6.0, 1.5, L2: $1 \times 10^{-4}$)}&\parbox{.14\textwidth}{\centering (10.0, 0.1, 1.8, 3.0, L2: $1 \times 10^{-5}$)}&\parbox{.14\textwidth}{\centering (50, 0.1, 2.5, 3.0, L2: $3 \times 10^{-5}$)}&\parbox{.14\textwidth}{\centering (50, 0.1, 2.5, 3.0, L1: $1 \times 10^{-3}$)}\\[3mm]

APL ($\alpha, \beta, \delta$)&(1.0, 1.0, L2: $1 \times 10^{-4}$)&(10, 0.1, L2: $1 \times 10^{-5}$)&(50, 0.1, L2: $3 \times 10^{-5}$)&(50, 0.1, L1: $1 \times 10^{-3}$)\\

ANL-CE ($\alpha, \beta, \delta$) &(5.0, 5.0, L1: $5 \times 10^{-5}$)&(10.0, 1.0, L1: $5 \times 10^{-7}$)&(20.0, 1.0, L1: $5 \times 10^{-6}$)&(5.0, 1.0, L1: $1 \times 10^{-3}$)\\

Drainage ($\alpha, \beta, \delta$) &(1.0, 1.0, L1: $2 \times 10^{-4}$)&(0.1, 10.0, L1: $1 \times 10^{-4}$)&(0.1, 10.0, L1: $1 \times 10^{-5}$)&(0.01, 100, L1: $1 \times 10^{-3}$)\\

\bottomrule
\end{tabular}
}
\caption{Parameters used per method on each dataset based on the results presented in the main manuscript. Due to the limitation, we have not fine-tuned SCE, APL and ALF losses on Clothing1M dataset. Aside from that, for Webvision and Clothing1M, the best parameters for all methods were obtained from \cite{ANL}.}
\label{table:parameters}
\end{table}

\begin{table*}[h!]
\centering
\begin{tabular}{lccccccccc}\toprule
& CE & GCE & SCE & AFL & APL & ANL CE &  Drainage & max.\ err.\\\midrule
CIFAR-10\\
~~~~~~noise $=0.0$ & 90.29 & 89.37 & 91.32 & 91.44 & 91.55 & 91.76   & 91.30 & $\pm$\textit{0.04} \\[1mm]
~~~\textit{asymmetric:}\\
~~~~~~noise $=0.2$ & 83.09 & 85.46 & 86.26 & 89.42 & 89.46  & 89.34   & 88.54 & $\pm$\textit{0.25}\\
~~~~~~noise $=0.3$ & 78.51 &79.83& 80.53& 86.50 & 86.96 & 85.13 & 87.56 & $\pm$\textit{0.32}\\
~~~~~~noise $=0.4$ & 73.38 & 72.78 & 73.59 & 78.65 & 80.90 & 78.07 & 84.66 & $\pm$\textit{0.39}\\
~~~~~~noise $=0.45$ & 70.79 & 69.91 &70.65 & 62.22 & 68.88 & 72.53 & 80.23 &$\pm$\textit{1.41}\\[1mm]
~~~\textit{instance dependent:}\\
~~~~~~noise $=0.2$ & 75.31 &  85.58 & 83.58 & 89.17 & 89.38 & 88.78 & 87.92 & $\pm$\textit{0.34}\\
~~~~~~noise $=0.4$ & 57.07 & 63.98& 63.86 & 77.33 & 77.68 & 76.54   &  81.95 &$\pm$\textit{0.90}\\
~~~~~~noise $=0.5$ & 46.48 & 50.87 & 50.66 & 47.06 & 49.34 & 58.41 & 64.22 &$\pm$\textit{
0.60}\\
~~~CIFAR10-N & 61.48 & 74.98 & 73.78 & 81.38 & 81.58 & 80.54 & 79.85 &$\pm$\textit{0.13}\\
\midrule
CIFAR-100\\
~~~~~~noise $=0.0$ & 70.11 & 62.10& 70.36&69.05& 65.60& 70.15 & 73.31 &$\pm$\textit{1.29}\\[1mm]
~~~\textit{asymmetric:}\\
~~~~~~noise $=0.2$ & 59.02 & 64.42 & 58.58& 63.55 & 60.56  & 66.09 & 70.04 & $\pm$\textit{0.87}\\
~~~~~~noise $=0.3$ & 50.48 & 62.16 & 50.27 & 56.77& 52.54 & 59.85 & 67.93 & $\pm$\textit{1.60}\\
~~~~~~noise $=0.4$ &41.79 & 54.44 & 41.48 & 45.21& 42.16 & 46.01 & 61.55 &$\pm$\textit{1.67}\\
~~~~~~noise $=0.45$ & 37.34 & 44.07 & 36.80 & 38.45 & 30.71 & 37.60 & 52.69 &$\pm$\textit{3.10}\\[1mm]
~~~\textit{instance dependent:}\\
~~~~~~noise $=0.2$ & 57.99 & 62.34 & 57.22 & 64.93 & 59.07 & 66.20 &67.16 &$\pm$\textit{1.23}\\
~~~~~~noise $=0.4$ & 43.19 & 60.69 & 43.31 & 52.57 & 38.31 & 56.55 & 60.63 &$\pm$\textit{1.20}\\
~~~~~~noise $=0.5$ & 35.31 & 47.21 & 34.61 & 41.81 & 22.55 & 45.67 &53.01 &$\pm$\textit{1.59}\\
~~~CIFAR100-N & 49.57 & 51.31 & 48.64 & 56.35 & 52.28 & 56.62 & 57.77 &$\pm$\textit{0.13}\\
\bottomrule
\end{tabular}
\caption{Ablation study, where we test the same methods and datasets as in Table \ref{table:synthetic} of the main paper, but where we change the regularization scheme. Specifically, CE, GCE, and SCE are this time computed using L2 regularization where as AFL, APL, ANL CE and Drainage are computed using L1 regularization. For each method, we observe a decrease in performance compared to the results reported in the main paper.}
\label{table:results_additional}
\end{table*}

\begin{figure*}[h!]
  \centering
      \includegraphics[width=.95\linewidth]{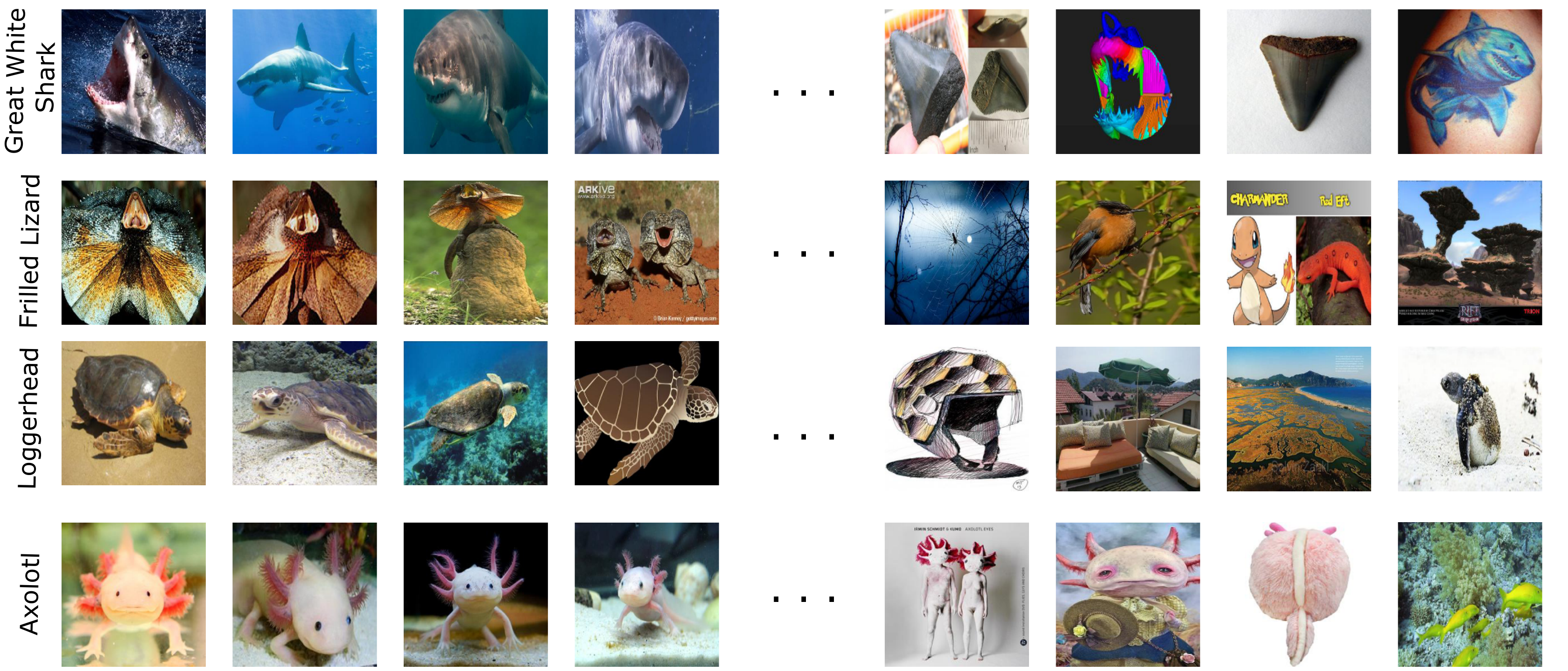}
  \caption{Same visualization as in Fig.\ \ref{fig:drain_images_webvision_validation} of the main paper, but where we consider this time the training samples ordered from lowest to highest drainage probability $p_d$. The training data has a higher fraction of noisy labels, and they predictably end up in the rightmost columns, i.e.\ with high drainage probability $p_d$.}
  \label{fig:drain_images_webvision_train}
\end{figure*}

\begin{figure*}[h!]
  \centering
      \includegraphics[width=.95\linewidth]{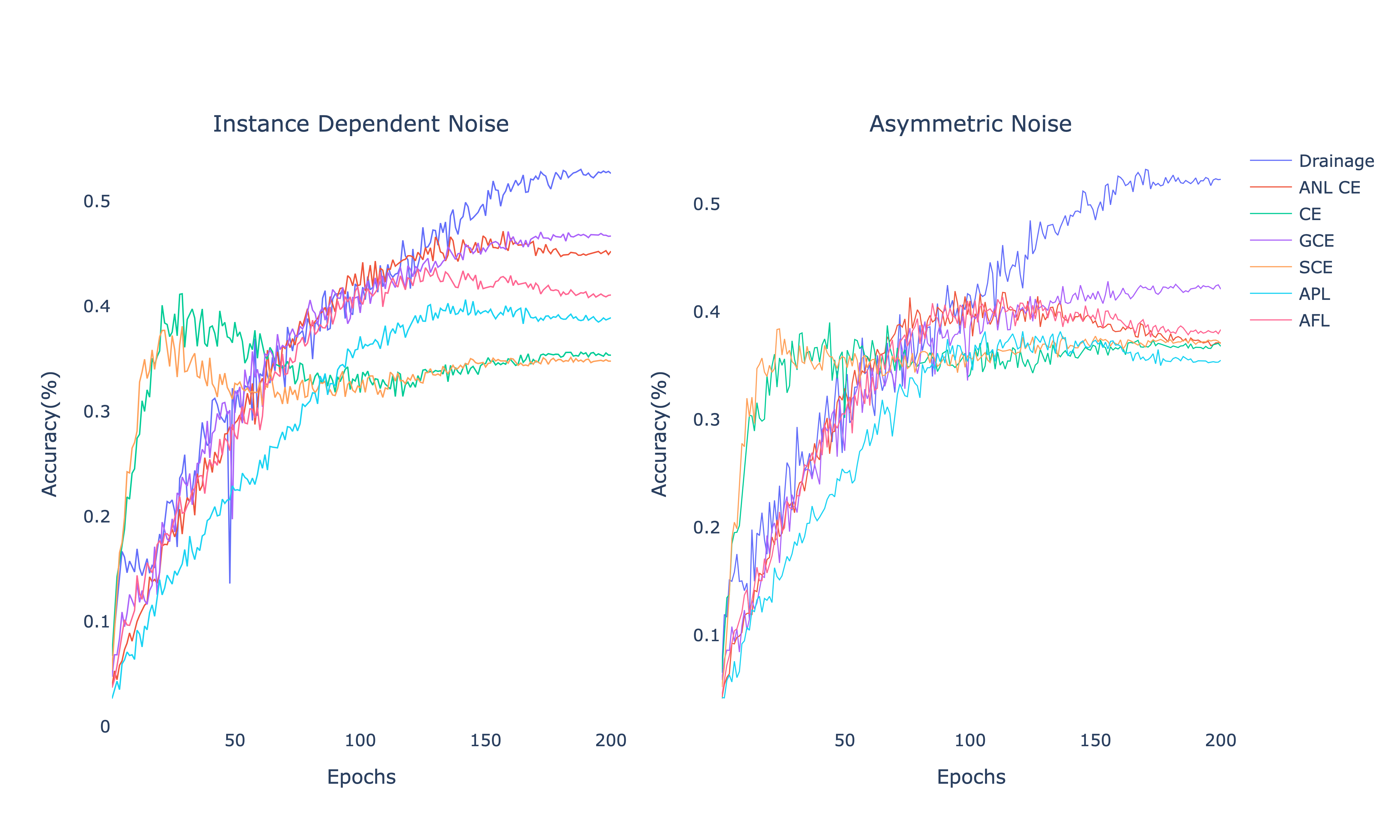}
  \caption{Test accuracies of loss functions on CIFAR-100 under instance dependent noise (50\%) and asymmetric noise (45\%). Paramters of loss functions are as based on Table ~\ref{table:parameters}.}
  \label{fig:training_curves}
\end{figure*}

\begin{table}[h!]
\centering
\begin{tabular}{lcc}\toprule
Loss function     & SVHN              & CIFAR10           \\\midrule
CE        & 94.6±0.9          & 80.6±8.9          \\
Drainage (\textbf{ours})  & \textbf{95.5±0.6} & \textbf{84.6±3.3}\\
\bottomrule
\end{tabular}
\caption{Results on Open Set Recognition: Closed Set Accuracy}
\label{table:osr-closed}
\end{table}

\end{document}